# Wavelet-Enhanced Desnowing: A Novel Single Image Restoration Approach for Traffic Surveillance under Adverse Weather Conditions

**Zihan Shen[1,2], Yu Xuan[3] and Qingyu Yang[1,2]**

[1] Automatic Control Research Institute, China Nuclear Power Engineering Co., Ltd., Beijing 100840, China
[2] CNNC Engineering Research Center for Fuel Reprocessing, Beijing 100840, China
[3] Chinese Aeronautical Establishment, Aviation Industry Corporation of China, Ltd, Beijing 100020, China

Corresponding author: Qingyu Yang (e-mail: yangqya@ cnpe.cc).

**ABSTRACT** Image restoration under adverse weather conditions refers to the process of removing degradation caused by weather particles while improving visual quality. Most existing deweathering methods rely on increasing the network scale and data volume to achieve better performance which requires more expensive computing power. Also, many methods lack generalization for specific applications. In the traffic surveillance screener, the main challenges are snow removal and veil effect elimination. In this paper, we propose a wavelet-enhanced snow removal method that use a Dual-Tree Complex Wavelet Transform feature enhancement module and a dynamic convolution acceleration module to address snow degradation in surveillance images. We also use a residual learning restoration module to remove veil effects caused by rain, snow, and fog. The proposed architecture extracts and analyzes information from snow-covered regions, significantly improving snow removal performance. And the residual learning restoration module removes veiling effects in images, enhancing clarity and detail. Experiments show that it performs better than some popular desnowing methods. Our approach also demonstrates effectiveness and accuracy when applied to real traffic surveillance images.

## I. INTRODUCTION

The quality degradation of images captured under adverse weather conditions, such as haze, snow, or rain, is primarily due to the presence of suspended weather particles in the air [1]. Veiling effect occurs when acquiring images. [2] (similar to the haze or the mist), while in heavy rainstorms or snowstorms, raindrops and snowflakes can even obscure targets entirely. In modern intelligent traffic surveillance systems, highly adaptive and robust image restoration algorithms can significantly enhance the performance of advanced computer vision tasks such as Object Detection [3] and Object Tracking [4].

To meet the practical requirements of image restoration in traffic surveillance, it is essential to acknowledge that image degradation in real-world monitoring scenarios demonstrates in a variety of complex and diverse patterns. Raindrops at different distances exhibits distinct visual characteristics in images: raindrops close to the camera appears as a veil mid-distance raindrops forms streaks and distant rain resembles fog. In traffic surveillance the outdoor characteristics of environments leads to unique challenges. Drizzle typically produces fog-like veiling effects, while pouring rain causes water droplets to adhere to the camera lens, resulting in complete image occlusion. It contrasts sharply with the dense, intricate rain streaks commonly represented in mainstream deraining datasets, such as Rain100H as shown in FIGURE 1 (b) [4]. Thus, the primary challenge in processing rainy traffic surveillance images lies in mitigating the veiling effect induced by fog-like rain.

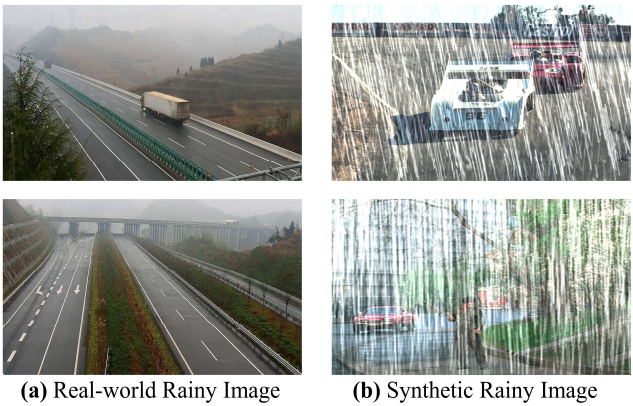

**(a) Real-world Rainy Image**     **(b) Synthetic Rainy Image**

**FIGURE 1. An example of real-world traffic surveillance image under rain conditions: (a) real-world rainy image captured in traffic camera, showing a fog-like veiling effect (b) synthetic rainy image, exhibiting dense, intricate rain streaks commonly found in Rain100H deraining datasets.**

Image degradation becomes worse in snowy conditions not only because heavy veiling effects but irregular shapes and larger surface area of snowflakes[5]. Due to fall slower than raindrops, snowy surveillance images are often populated by a combination of snowflakes, snow streaks, and veiling effects. Hence de-veiling and desnowing are two main challenges when restoring traffic surveillance images effectivity.

Recently, data-driven deep learning methods have demonstrated remarkable performance in weather-related image restoration. For instance, DehazeNet effectively removes veiling effects by learning the mapping between hazy images and medium transmission maps [7], while DesnowNet was a pioneering application of CNNs to desnowing tasks [8]. The introduction of the SRRS dataset facilitated systematic studies of veiling effects, and wavelet decomposition marked its first application in the desnowing domain [9]. Over time, increasingly advanced deweathering architectures have emerged, incorporating innovations such as refined structural designs [10], [11], [12], [13], visual prompting [14], [15], mixtures of experts (MoE)[16], and multimodal models[17], [18], [19]. However, the computing power requirements of Large Model make them unsuitable for intelligent surveillance systems, which operate in proprietary environments and lack hardware upgrades or replacement. Moreover, existing methods often struggle to capture the specific characteristics of snowy scenes in real-world surveillance, making it challenging to meet the actual deweathering requirements for surveillance image restoration.

In this paper, we propose a novel image restoration architecture based on Dual-tree Complex Wavelet Transforms (DTCWT) to address the unique degradation patterns encountered in traffic surveillance images under adverse weather conditions. Our method incorporates a DTCWT-based snow feature enhancement module and a Dynamic Convolution Acceleration module to mitigate snow-related degradation. Additionally, a Residual Learning Restoration module is employed to address veiling effects caused by rain, snow, and fog, resulting in a unified approach that improves the efficiency and reliability of intelligent surveillance systems under adverse weather conditions. This paper is the first to apply deep learning-based snow removal techniques to real-world surveillance image restoration tasks.

The main contributions in this paper are summarized as follows:

- Precise snow region analysis: By leveraging the decomposition properties of DTCWT, the proposed network architecture accurately extracts and analyzes information from snow-covered regions, significantly improving the performance of snow removal.
- Residual learning for de-veiling: The residual learning restoration module effectively removes veiling effects in images, enhancing clarity and detail.
- Dynamic convolution for efficient desnowing: To address the dual challenges of computational redundancy in multi-layer DTCWT decompositions and suboptimal training convergence, we introduce dynamic convolution operators into the desnowing framework [46].

The paper is structured as follows: Section II offers a review of existing image desnowing approaches and surveillance image restoration under adverse weather. The methodology of proposed model is presented in Section III. Section IV discusses and analyze the results of experiment. Section V presents the conclusions.

## II. RELATED WORK

### A. SINGLE IMAGE DESNOWING
With the rapid advancements in artificial intelligence, methods based on deep learning have achieved significant success in snow removal tasks. The pioneering deep neural network for snow removal, DesnowNet, utilized a multi-stage model to progressively eliminate snow particles and flakes[8]. [10] proposed the All-in-One Adverse Weather Removal Network, which incorporates multiple specialized encoders to simultaneously address various types of degradation phenomena, enabling the restoration of images under diverse adverse weather conditions. [9] have focused on image snow removal research for years, developed JSTASR, a method that considers both veiling effects and the diversity of snow. Additionally, [20] were the first to integrate the Dual-Tree Complex Wavelet Transform (DTCWT) into a network architecture and designed a prior-based contradictory channel loss function tailored for snow removal tasks. [21] proposed TransWeather, which utilizes Vision Transformer (ViT) as the backbone network and introduces a weather-type decoder to effectively handle various adverse weather scenarios, achieving remarkable performance improvements. [22] explored deep invertible models to separate snow-covered components. [23] introduced SmartAssign, a method for learning intelligent knowledge assignment strategies to tackle both de-raining and snow removal tasks. [24] proposed MspFormer, a Transformer-based model with a multi-scale projection mechanism for enhanced snow removal. [25] introduced HCSD-Net, leveraging color space transformation to extract snow cover features for precise and efficient snow removal. Compared to traditional single-snow removal methods, we highlight the limitations of latter to address real-world complexities and emphasize AiOIR's significant advantages in efficiency, adaptability, and scalability [51]. Current models still struggle to handle complex and composite degradations, lack computational efficiency, and do not generalize well in real-world scenarios.

### B. WAVELET TRANSFORM FOR IMAGE APPLICATIONS
Wavelet transforms have demonstrated significant potential in addressing various low-level image processing tasks. In image denoising, [26] use multivariate statistical methods to estimate wavelet transform coefficients for noise suppression. [27] enhance denoising by integrating wavelet transforms with genetic algorithms. [28] utilize wavelet transforms in combination with CNNs for image super-resolution, leveraging their capability to capture detailed and contextual information. [29] improve super-resolution tasks by employing cross-connections and residual networks to merge CNNs with wavelet transforms.

The DTCWT [41], a powerful extension of wavelet families, has seen diverse applications in image processing. [30] apply DTCWT and scattering networks for image classification. [31] and [32] utilize DTCWT for image segmentation tasks. [33] employ DTCWT in 3D image registration, leveraging its phase-shift properties, while [34] and [35] integrate it with adaptive histogram equalization for low-light image enhancement.

As evidenced by the forementioned studies, wavelet transforms have proven effective for low-level image processing tasks. Inspired by these applications, we introduce two cascaded wavelet transform techniques into CNNs to fuse frequency and structural information, aiming to enhance the visual performance of snow removal tasks.

### C. SURVEILLANCE IMAGE RESTORATION
Surveillance image restoration is a primary application area for adverse weather image restoration algorithms. In recent years, an increasing number of studies have applied rain and haze removal techniques to traffic surveillance images. For instance, [36] utilized contrastive learning with weakly paired images for rain removal in traffic scenes, while [37] employed tensor low-rank minimization to effectively remove rain and retrieve clean backgrounds. [38] leveraged the visual characteristics of rain streaks for multi-frame rain removal in surveillance images, adaptively extracting features from varying resolutions and frame rates. Similarly, [39] developed lightweight adversarial learning networks incorporating multi-resolution analysis and multi-scale encoders to address rain and haze removal in surveillance imagery. Furthermore, [40] utilized real-world all-weather images, synthesized images with enhanced weather noise, and integrated object detection with adverse weather image denoising. However, these methods fail to address the central features of degradation in traffic surveillance images, which often necessitates a focused approach toward snow removal.

## III. PROPOSED METHOD

### A. EFFICIENT REPRESENTATION OF SNOW BASED ON DTCWT
Dual-Tree Complex Wavelet Transform (DTCWT) represents a more advanced technique for image and signal processing than Discrete Wavelet Transform (DWT) [42]. While DWT is widely used for multi-scale image and signal processing with low redundancy, it has key limitations. First, noise sensitivity affects recovery accuracy, particularly in high-frequency components. Second, its weak directional selectivity limits performance. The standard DWT lacks good directional selectivity when describing specific directional features in images, such as edges, ridgelines, and diagonals, making it difficult to locate and separate details in different directions [43]. The DTCWT addresses these issues by using a pair of complex-valued wavelet filters to create a filter tree structure which enables multi-angle, multi-level, and multi-directional signal analysis. Unlike single-path decomposition, DTCWT uses a dual-tree structure in the complex domain with a complex wavelet basis. This design improves noise tolerance. The dual-tree structure enhances local statistical properties, suppressing noise more effectively, especially in high-frequency details, and reducing noise interference with useful information. Additionally, DTCWT improves directional selectivity. With complex wavelet filters, it performs decomposition along multiple orthogonal directions, accurately capturing and extracting geometric features in images regardless of their orientation [44], [45].

$$\psi(x) = \psi_h(x) + j\psi_g(x) \tag{1}$$

Where $j$ represents the imaginary unit, $\psi_h, \psi_g$ corresponds to two wavelet bases, representing the real and imaginary parts of the mother wavelet. These two basis functions combine to form wavelet bases with strong directional and frequency selectivity in two-dimensional space. The two-dimensional DTCWT decomposition process produces seven distinct frequency bands. One low-pass subband preserves the overall structure and smooth regions of the image, while the other six high-pass subbands capture high-frequency information corresponding to specific directions: $\pm15°$, $\pm45°$, and $\pm75°$. This enhanced selectivity makes DTCWT effective for extracting directional textures and edge details [41].

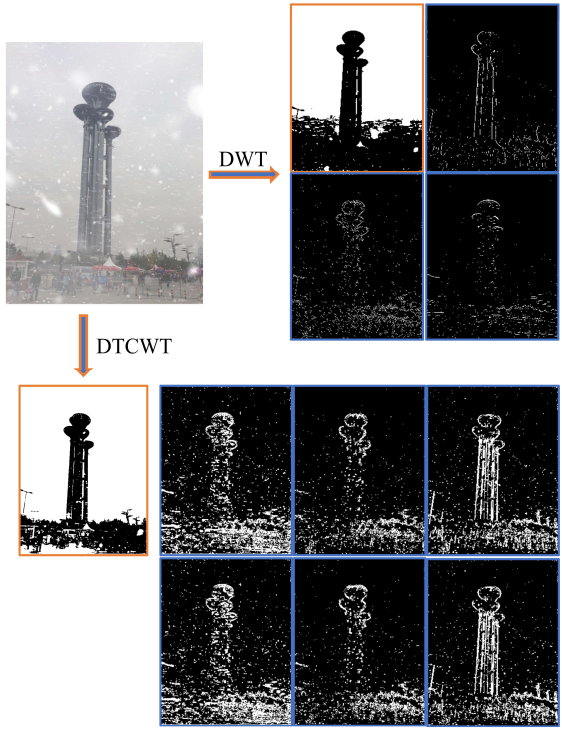

In snow images, snowflakes and snow streaks typically are slanted, directional structures with distinct textures. These structures have clear directional features and fine details. The effectiveness of extracting snow particle information using DTCWT is shown in FIGURE 2. It is obvious that the high-frequency subbands of DTCWT can effectively capture these features, particularly excelling in representing detailed structures like snow streaks [20]. By integrating DTCWT into deep learning network architectures and leveraging its unique properties, snow-covered region information can be more accurately extracted and analyzed. This integration significantly enhances the performance and precision of tasks such as snow cover detection and snow depth estimation.

### B. DTCWT-BASED IMAGE DESNOWING

#### 1) FRAMEWORK ARCHITECTURE

The overall network architecture is illustrated in FIGURE 3, presenting a composite, multi-layer deep neural network model comprised of three key components. The model begins with a five-layer Dynamic Convolution Acceleration (DCA) module, followed by two eight-layer Dual-Tree Complex Wavelet Enhancement (DTCWE) modules, and concludes with a ten-layer Residual Learning Restoration (RLR) module to refine the final image restoration. The DCA module's internal structure is as follows: the initial layer employs a 5×5 convolutional kernel for downsampling the input image, producing an output channel size of 64. Layers two through four are composed of one or two Weight Generators (WG) and a 5×5 convolutional layer. The fifth layer uses a ReLU activation function to extract and combine dynamic convolution features, keeping the output at 64 channels. Each WG includes a pooling layer, a 1×1 convolutional layer, a nonlinear activation function, and a Softmax classifier. The dynamic convolution mechanism calculates four weight coefficients using the WG, adjusting the parameters of four parallel convolutional kernels. This creates adaptive weighted convolution applied to the output from the first layer.

The output of DCA feeds into the DTCWE module. DTCWE utilizes the DTCWT to convert the original linear image structural information into seven frequency-domain feature components, including one low-frequency feature and six high-frequency features. The high-frequency features focus on capturing fine textures and directional textures of the image, such as streaks in snow scenes. These frequency features are optimized through a feature enhancement mechanism, which uses a four-layer residual dense connection, with each set of three layers consisting of a 5×5 convolution layer and activation function. The final layer employs a 1×1 convolution layer, aimed at suppressing adverse factors such as snow streaks, while restoring more detailed information from the image. Finally, the enhanced frequency-domain features are mapped back to the linear space using the Inverse Dual-Tree Complex Wavelet Transform (IDTCWT) without loss of information.

The RLR module further refines the information from the DTCWT. It applies residual learning twice, with each process containing four residual dense layers and ReLU activation functions. The network progressively refines the image to remove the veil effect. Each stage uses a 5×5 convolution kernel to gradually suppress noise and enhance features as the network deepens. The outputs of the two residual blocks are deeply fused and passed through another convolution layer with a ReLU activation function for final feature fine-tuning.

#### 2) DYNAMIC CONVOLUTION ACCELERATION MODULE

CNNs in image restoration tasks utilize a weight-sharing mechanism in convolutional layers, ensuring that the same weights are applied uniformly across all input images within the same layer. While this simplifies the model, it limits the network's ability to dynamically adjust parameters based on image content, thereby affecting the robustness of the classifier. Inspired by research in image classification and denoising, the Dynamic Convolution [46] is introduced into the snow removal task to enhance training efficiency and address the increased model complexity caused by the DTCWT.

In common perception models, $W$, $b$, and $g$ represent the weight, bias, and activation function, respectively, as shown in (2).

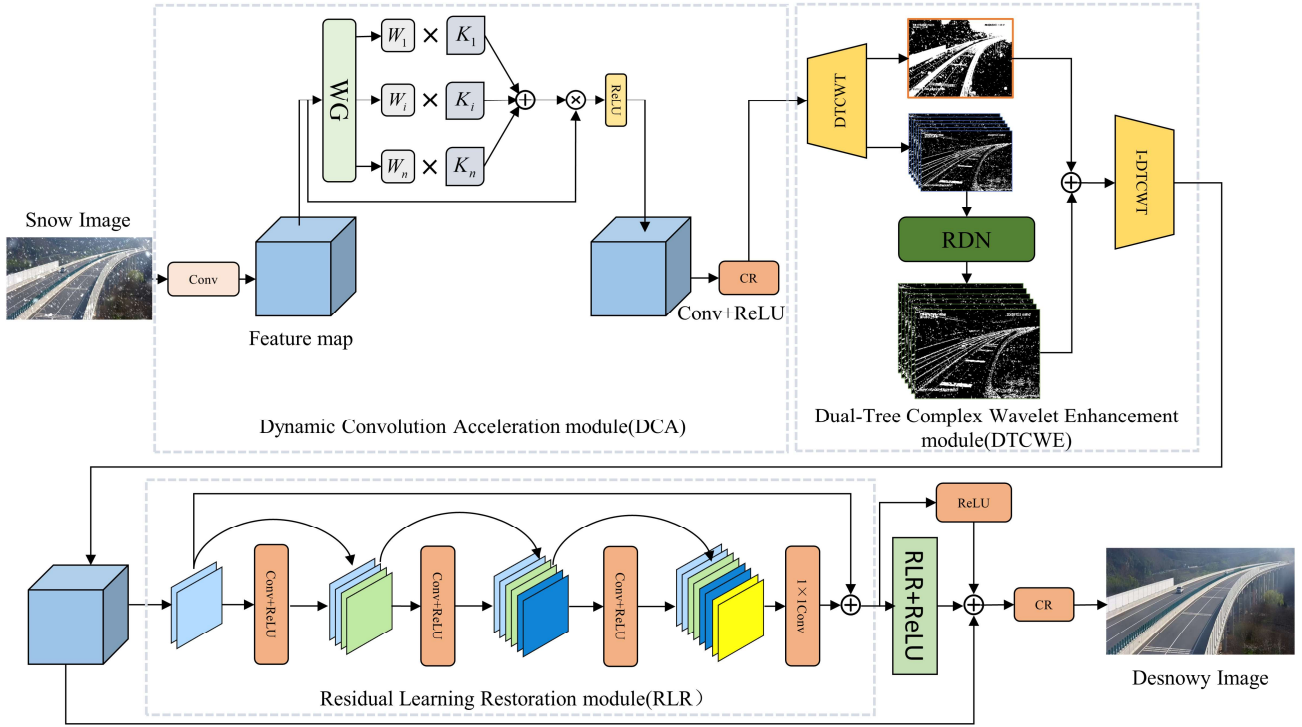

**FIGURE 3.** Overview of the proposed wavelet-enhanced desnowing framework, showing a multi-layer neural network for image restoration, comprising a DCA module, two DTCWE modules, and an RLR module.

$$y = g(W^T x + b) \qquad (2)$$

Dynamic convolution, involves the use of weight coefficients to compute a weighted sum of weights and biases, with these coefficients adapting to changes in the input data. This process can be expressed by (3).

$$y = g(\tilde{W}^T(x)x + \tilde{b}(x))$$
$$\tilde{W}^T(x) = \sum_{i=1}^{N} \pi_i(x)\tilde{W}_i \qquad (3)$$
$$\tilde{b}(x) = \sum_{i=1}^{N} \pi_i(x)\tilde{b}_i$$

Where $\pi_i(x)$ represents the weight coefficient, which depends on the input information $x$, meaning that its value varies with the input. Therefore, for each specific input image, DCA can represent the image information in best linear combination way. Furthermore, considering the nonlinear nature of the model, DCA exhibits a powerful representational ability beyond that of linear models, enabling more flexible and precise capture and modeling of the potential complex structures and diverse features within the input data.

The WG produces four weights, which are then applied in a weighted manner to four parallel convolutional kernels, allowing for dynamic adjustment of the convolution parameters. Its main structure is shown in FIGURE 4. The WG module dynamically generates adaptive weights based on the content of the input image. The results from the four weighted convolutional kernels are then combined through a 5×5 convolution layer, with both the input and output channels set to 64. The convolution operation performs a per-channel convolution between the output of the first convolutional layer from the WG module and the second output from the dynamic convolution process. This step aims to merge features from both sources and adaptively determine the most suitable parameter settings for different snow-degraded images. In this process, $WG_i$ and $K_i$ represent the i-th weight parameter generated by the WG module and the k-th convolution kernel among the four parallel convolutional kernels, respectively. Additionally, Average Pooling (AP) and Softmax operations are incorporated into the model for further feature dimensionality reduction and probability calculation. The detailed structure of the WG module includes a first convolutional layer with 64 input channels and 4 output channels, and a second convolutional layer with 4 input and output channels.

### 3) DUAL-TREE COMPLEX WAVELET ENHANCEMENT MODULE

The DTCWE module presented in this section utilizes the multi-scale characteristics of the DTCWT to decompose and process snow-degraded surveillance images. After processing through the dynamic convolution module, these images are further decomposed by DTCWT into six high-frequency components and one low-frequency component. The high-frequency components capture edge and detail information, which, in the context of snow images, correspond to the fine structures of snow particles. Meanwhile, the low-frequency component captures the overall structure and broader color regions, representing larger snow streaks or extensive snow-covered areas. During processing, this low-frequency component is recursively decomposed into six high-frequency subbands, each representing distinct directional features. To further enhance the results, a deep residual dense enhancement network is applied to suppress residual snow streaks across various scales and restore fine image details.

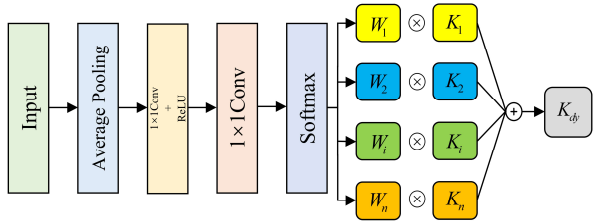

**FIGURE 4.** Architecture of Weight Generator (WG).

The Residual Dense Network (RDN) [47] is a residual learning architecture designed to effectively extract local features from images or signals using densely connected convolutional layers. In the RDN, each layer's output is not only passed forward to subsequent layers but also back-propagated to all following layers, forming what is known as a Continuous Memory (CM) mechanism. This design ensures the comprehensive flow and accumulation of information, enabling the network to fully exploit previously learned features, prevent gradient vanishing or explosion issues, and enhance its overall learning capacity. In the DTCWE, the RDN is employed as the feature enhancement network for each image component after DTCWT decomposition. As illustrated in FIGURE 2, snow-degraded images decomposed via DTCWT produce frequency components that capture directional snow streaks. By leveraging the RDN for feature extraction and snow streak suppression, the model can adaptively learn the local features of each component, generating more effective and meaningful feature representations. After the RDN successfully extracts a comprehensive set of local features, the next step is global feature fusion—a process that integrates high-level features. This step consolidates the diverse and detailed features extracted by the local Residual Dense Blocks (RDBs) into a unified enhancement, offering a global perspective. By adaptively learning the hierarchical global features of the image, the model significantly improves its ability to identify and suppress snow particles and streaks, leading to more accurate restoration outcomes.

The feature enhancement module utilizes four residual dense layers to optimize extracted features. As depicted in FIGURE 5, the first three layers employ a 5×5 convolution kernel to perform convolution operations, effectively capturing spatial correlations and extracting refined and abstract feature representations. Each of these layers is followed by a ReLU activation function, which introduces non-linearity, enabling the model to learn more complex patterns, mitigate the gradient vanishing issue, and enhance deep feature learning. After processing through these densely connected convolution layers, the final layer incorporates a

1×1 convolution kernel. While this smaller kernel does not expand the spatial receptive field, it compresses and fuses deep features in the channel dimension, thereby strengthening feature representation and improving the model's generalization capability. This architecture ensures a seamless and efficient transformation of the initial frequency features into optimized features, significantly enhancing the model's expressiveness and accuracy in handling the task.

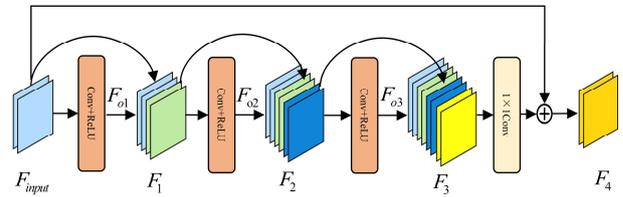

**FIGURE 5.** Architecture of Residual Dense Net (RDN).

The feature enhancement applied to low-frequency components improves the model's ability to recover from snow veil effects and large-scale snow particles, while enhancement on high-frequency components strengthens its capability to address snow streaks and other directional snow degradation effects. Collectively, this feature enhancement mechanism refines and amplifies the original image features, thereby enhancing the model's ability to discriminate and address snow-related degradation. Following processing at each scale, the enhanced high-frequency and low-frequency subbands are merged using the IDTCWT to reconstruct the image. This method enables effective feature enhancement for snow images across multiple scales, particularly in removing larger snow streaks and intricate snowflake patterns. Furthermore, the wavelet transform's superior geometric representation across various directions ensures accurate capture and processing of snow information in different frequency subbands, facilitating effective removal of snow artifacts regardless of particle size or shape. After undergoing DTCWE processing, the primary remaining degradation in the image is attributed to the veil effect, which causes blurring. This is effectively addressed by the ten-layer residual learning restoration module, resulting in a clear and clean restored image.

### 4) RESIDUAL LEARNING RESTORATION MODULE

The Residual Learning Network [48] has been integrated into the feature enhancement mechanism of DTCWE. To differentiate the RDN in the DTCWE module from the residual learning module designed for addressing veil effects, the latter is referred to as the RLR module. The RLR focuses on eliminating degradation caused by rain, snow, and haze veil effects in surveillance images. The RLR module features a 10-layer architecture that incorporates a dual mechanism combining the RDN and ReLU activation functions. It begins with four RDN layers paired with ReLU activation functions, forming the first-level feature refinement unit. This structure is designed to extract detailed feature representations essential for effectively removing veil effects.

Following this, two additional RDBs are introduced, integrating feature information generated by the second convolution layer and ReLU activation functions. This configuration addresses long-range dependency issues, facilitating the efficient transfer of shallow image features to deeper layers. To mitigate overfitting during feature enhancement, convolution layers with consistent input and output channels (both set to 64) are used in conjunction with ReLU activation functions for fine-tuning and refining previously generated features.

Finally, the degradation map is reconstructed using convolution layers and merged with the original image via residual learning to produce a clean output. In this process, all convolution kernels in the residual blocks are set to 5×5. The two RDN modules have fixed input and output channels of 64, while the final convolution layer processes 64 input channels to generate a 3-channel restored surveillance image.

## C. LOSS FUNCTIONS
With the development of single-image snow removal tasks, many studies have fully considered the characteristics of snowy images and, combining advanced research, proposed a new operation that can identify the differences between snowy and non-snowy images, known as the Contradict Channel (CC) [20]. The introduction of Dark Channel Prior (DCP) (4)has played a significant role in advancing research and development in the field of image dehazing [49].

$$I_{\text{Dark}}(x) = \min_{y \in \Omega(x)} \left( \min_{c \in \{r,g,b\}} I^c(x) \right) \quad (4)$$

This pioneering concept effectively utilizes the inherent statistical properties of natural images, and through a deep insight into the commonly observed dark channel phenomenon in haze-free images, it constructs a powerful inverse estimation method for image degradation models. An increasing number of studies aim to discover the unique characteristics of images in severe weather conditions, leading to the development of Contradict Channel Prior (CCP). Its expression is shown in (5).

$$I_{\text{Contradict}}(x) = \max_{y \in \Omega(x)} \left( \min_{c \in \{r,g,b\}} I^c(x) \right) \quad (5)$$

In the image color channel $c$, $I^c(x)$ represents the intensity value at the pixel location, while $\Omega(x)$ represents a local image patch centered at $x$ with a fixed size. The CCP operation is applied to both snowy and clear images. Compared to clear images, snowy images show a significant increase in the number of pixels close to 1 after the CCP operation. The intensity in snow-covered regions is higher than in non-snow regions. This phenomenon is mainly attributed to three factors: first, the characteristics of snowflakes themselves; second, the presence of snow stripes; and third, the veil effect caused by snow. These factors contribute to the higher intensity of the contradict channel. On the other hand, compared to snowy images, clear images often contain brightly colored objects, such as blue waters, green vegetation, and red or yellow flowers. These objects can lead to a relatively lower intensity in the contradict channel. The expression for the image brightness channel operation is shown in (6).

$$I_{\text{Bright}}(x) = \max_{y \in \Omega(x)} \left( \max_{c \in \{r,g,b\}} I^c(x) \right) \quad (6)$$

FIGURE 6 illustrates the dark channel, brightness channel, and contradict channel for snowy images, along with their corresponding clean image comparisons. It is evident that while the dark channel can capture some of the veil effect in snowy images, its reliance on the minimum pixel value in local regions limits its effectiveness in detecting and extracting smaller snowflakes. Similarly, the brightness channel partially reflects the distribution of snow. However, due to its reliance on selecting the maximum value from each color channel, it may fail to capture snow features accurately, as the saturation of a single color can result in balanced color representation in non-snowy regions. This reduces the brightness channel's ability to distinguish the unique characteristics of snowy images.

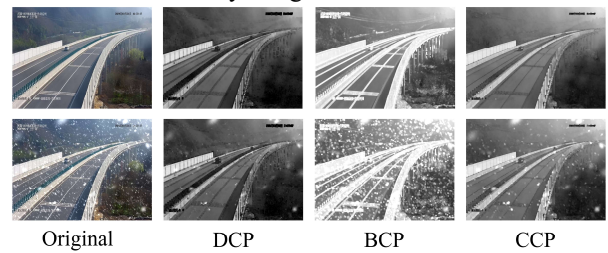

| Original | DCP | BCP | CCP |

**FIGURE 6.** Contrast of the Contradiction Channel in Snowy Image.

In contrast, the contradict channel demonstrates superior discriminative capability in identifying snow-covered scenes. The intensity differences in this channel serve as a natural quantitative metric to differentiate between snow-covered and non-snow images. Leveraging this property, a Contradict Channel Loss (CCL) is proposed to provide a more targeted optimization strategy during the training of the snow removal algorithm.

$$L_{CC} = \| CC(y) - CC(x) \| \quad (7)$$

The CCL is mathematically defined as (7). Here, $CC$ represents the contradict channel operation, $L_{cc}$ denotes the contradict channel loss, $x$ refers to the clean image, and $y$ represents the image restored by the algorithm model.

## IV. EXPERIMENTS

### A. DATA AND IMPLEMENTATION DETAIL
The experiments were conducted using the PyTorch deep learning framework and Python 3.8 as the programming language. The GPU used was the NVIDIA GeForce RTX 4090:24GB, featuring a GDDR6X memory configuration and 168 RT Cores The network training batch size was set to 16, and the Adam [50] optimizer was applied. The initial learning rate was set to $10^{-4}$, and training ran for 100 epochs.

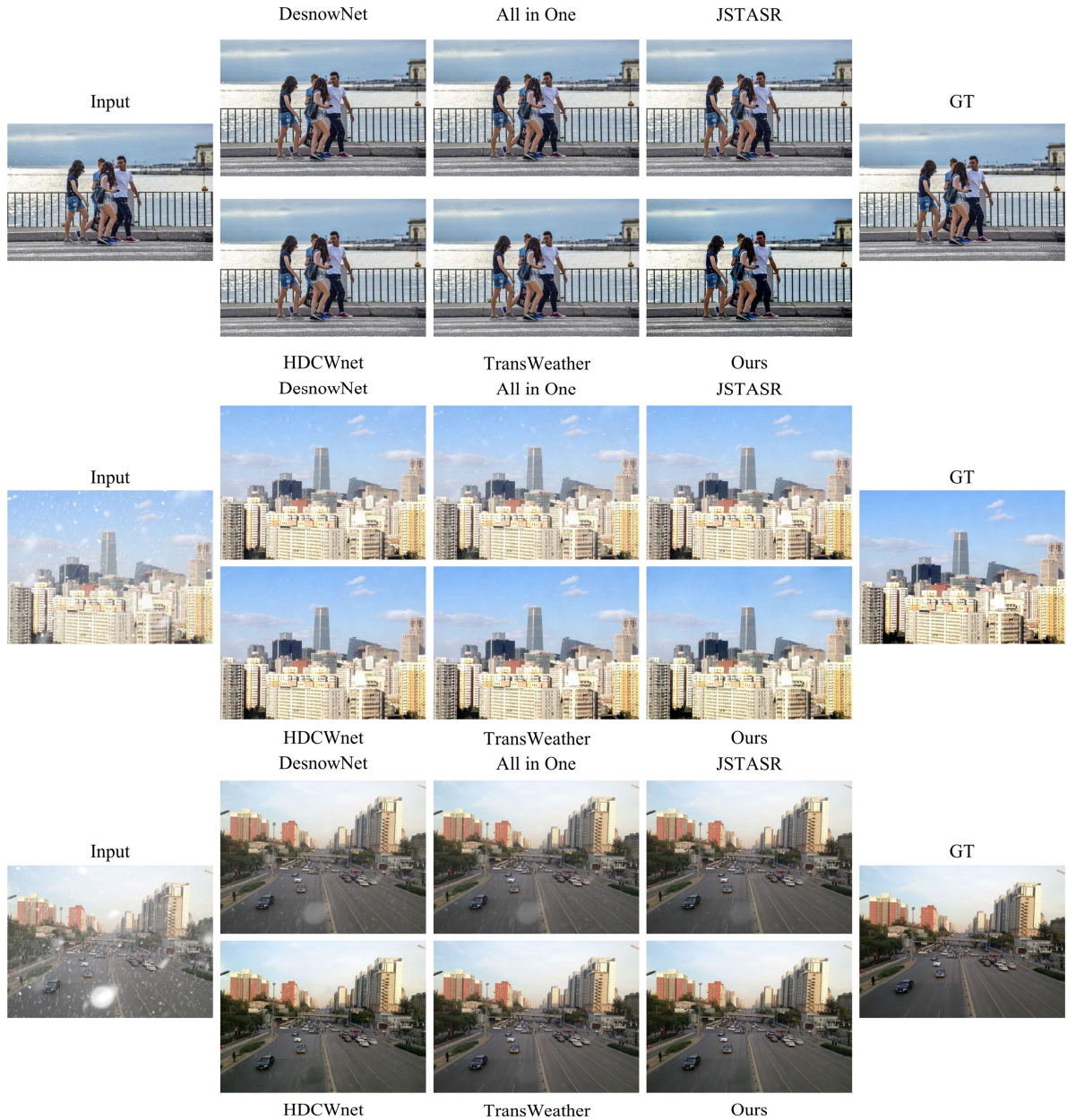

The learning rate varied as follows: from epoch 1 to 30, it was $10^{-4}$; from epoch 31 to 60, it was $10^{-5}$; and from epoch 61 to 100, it was $10^{-6}$.

We designed a model to remove snow and veiling effects, including those caused by rain and fog, focusing on restoring highway surveillance images. To evaluate our method's generalization, we used a large training dataset from mainstream sources and tested it on three public datasets: Snow-100K[8], SRRS [9], and CSD [20]. The Snow-100K dataset comprises 100,000 synthetic snow images paired with their corresponding clear images, along with 1,329 real snow images. The SRRS dataset includes 15,000 synthetic snow images and 1,000 real snow images collected from internet sources. The CSD contains 10,000 synthetic snow images, where snowflakes and snow streaks are synthesized with varying attributes, such as transparency, size, and position. Additionally, Gaussian blur is applied to the snow particles in CSD to better simulate real-world snow scenes.

## B. EXPERIMENT RESULT AND ANALYSIS

We trained the network model designed in this paper using the CSD dataset on the platform outlined earlier. The CSD dataset includes snow particles, snow streaks, and simulated veiling effects, making it highly representative of the key characteristics of traffic surveillance images The training parameters are configured according to the experimental setup described in Section A. Subsequently, we perform a comparative analysis using the trained network model. The Snow-100K dataset is recognized as a benchmark for snow removal tasks in the community. In this section, we evaluate DesnowNet, All in One, JSTASR, HDCWnet, TransWeather, and the proposed network model (referred to as "Ours" in TABLE II) on the Snow-100K dataset. The test results are presented in TABLE II.

TABLE I
COMPARISON OF DIFFERENT NETWORKS IN SRRS AND CSD DATASET

| Dataset | Methods (PSNR(↑)/ SSIM(↑)) | | | | | |
|---------|-----------|------------|--------|--------|--------------|------------|
|         | DesnowNet | All in One | JSTASR | HDCWnet | TransWeather | Ours |
| SRRS | 20.38/0.84 | 24.98/0.88 | 25.82/0.89 | 27.78/0.89 | **28.03/0.90** | 27.52/0.88 |
| CSD | 20.13/0.81 | 26.31/0.87 | 27.96/0.88 | 29.06/0.90 | 29.32/0.90 | **29.89/0.91** |

TABLE II
COMPARISON OF DIFFERENT NETWORKS IN SNOW-100K

| Method | PSNR(↑) | SSIM(↑) |
|--------|---------|---------|
| DesnowNet(TIP2018) | 27.17 | 0.8983 |
| All in One(CVPR2020) | 28.33 | 0.8820 |
| JSTASR(ECCV2020) | 25.32 | 0.8076 |
| HDCWnet(ICCV2021) | 28.18 | 0.8800 |
| TransWeather(CVPR2023) | **28.48** | 0.8908 |
| Ours | 28.39 | **0.9021** |

As shown in TABLE I, the proposed network achieves the third-highest PSNR score while ranking first in the SSIM metric, surpassing TransWeather by a margin of less than 0.01. The first group in FIGURE 7 illustrates sample test results using the proposed network on the Snow-100K dataset. A closer examination of the third group of images reveals that the proposed network tends to produce darker overall tones. While it effectively removes small snow particles, some overfitting is evident in the image restoration process.

This outcome can be attributed to the network's design, particularly in the third stage, which prioritizes mitigating veiling effects caused by severe weather conditions. The RLR module, specifically designed to address veiling effects, contributes to overfitting when applied to the Snow-100K dataset, which lacks veiling effects. Consequently, the restored images appear darker overall. However, this design also has a positive impact significantly enhancing the restoration quality of background areas. Given that background regions constitute a substantial portion of the images, their improved restoration contributes notably to the structural similarity evaluation, resulting in a higher SSIM score.

The data in TABLE I indicate that the proposed network performs well on the CSD dataset, which includes added veiling effects. The PSNR score is 0.57 higher than TransWeather, and the SSIM score is 0.02 higher. Compared to other classical snow removal networks, the proposed network achieved better results. Through data analysis, it can be concluded that the proposed network has made significant improvements in these metrics. This improvement may be partly attributed to the fact that the network's training dataset, CSD, includes snow-adding patterns, which likely introduce some inertia in performance. Moreover, these results suggest that the proposed RLR module, designed to mitigate veiling effects, plays a crucial role in restoring real surveillance images. Additional ablation experiments will be conducted to further validate this hypothesis.

To illustrate the differences in snow removal performance across various networks, we present the restored images in the second and third groups of FIGURE 7 after applying these networks to the CSD test set. Upon closer inspection, several common trends emerge. Notably, we observe that all networks effectively remove snow from the sky background. This can be attributed to the fact that the sky contains less information, making it easier to restore, and its color closely matches that of the snowflakes, limiting the visible differences to the naked eye.

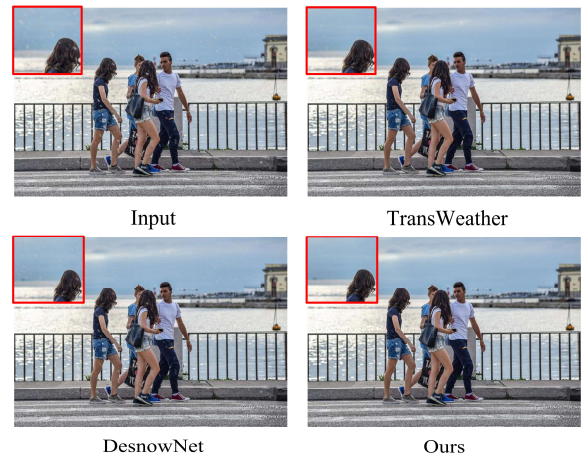

Input                    TransWeather

DesnowNet                Ours

**FIGURE 8.** Comparison of detail restoration using different algorithms.

When we examine the restoration of the road surface in FIGURE 7, it becomes clear that none of the networks entirely remove the snow obstructing the road, especially when large snowflakes are present. However, our proposed algorithm performs better than the others, although we note that color restoration could still be improved. The details of image restoration are illustrated in FIGURE 8. As can be seen from the figure, the algorithm proposed in this paper demonstrates a significant advantage in handling background snowflakes compared to other algorithms. To evaluate the practical effectiveness of the dynamic convolution module, the DTCWE module, and the residual restoration in the third stage, we conducted a series of ablation experiments. We then compared and analyzed the performance metrics, which are shown in TABLE III. A detailed review of these metrics reveals that both the dynamic convolution module and the DTCWE module significantly enhance the restoration quality. Furthermore, the ablation tests of the RLR module confirm its ability to remove the veiling effect. These results strongly support the practical effectiveness of the architecture we propose for real-world applications.

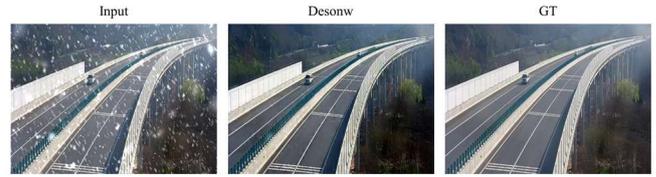

**FIGURE 9.** Traffic surveillance image restoration effect.

TABLE III
COMPARISON OF METRICS FOR NETWORK ABLATION EXPERIMENTS

| Network | PSNR(↑) | SSIM(↑) |
|---|---|---|
| Ours | 29.89 | 0.92 |
| Without DCA | 28.33 | 0.90 |
| Without DTCWE | 27.6 | 0.85 |
| Without RLR | 28.45 | 0.81 |

The time cost (Times) and the number of required parameters (Parameters) are shown in TABLE IV. The number of parameters refers to the total count of learnable parameters in the model, which includes all the weights and biases in the network. These parameters are variables that we update during training through backpropagation and optimization algorithms, and they govern how the network transforms, abstracts, and makes decisions on the input data. We compare the proposed method with other deep learning-based snow removal algorithms, and the results clearly show that our method requires significantly fewer computational resources. It reduces the running time by at least 74% and the number of parameters by 55%, respectively.

TABLE IV
COMPARISON OF RUN TIME AND THE NUMBER OF PARAMETERS

| | DesnowNet | JSTASR | Ours |
|---|---|---|---|
| Times(s) | 1.38 | 0.87 | 0.23 |
| Parameters | $1.57 \times 10^7$ | $6.5 \times 10^7$ | $6.7 \times 10^6$ |

In the final part of this section, we conduct empirical evaluations to validate the practical effectiveness of the proposed network architecture in traffic surveillance image restoration tasks. As shown in FIGURE 9, we apply the designed network to highway monitoring image restoration, achieving a PSNR of 24.67 and an SSIM of 0.81. Upon visual inspection, we observe that the overall restoration effect is quite good, although some minor imperfections remain in certain details. Our analysis suggests that highway monitoring images typically focus on the road surface, while the sky background—which is easier to restore—occupies less space. The road surface often contains more occluded details to restore harder.

## V. CONCLUSION

In this paper, we proposed a novel desnowing method for traffic surveillance image. Firstly, by analyzing the quality degradation of highway surveillance images in adverse weather, it clarifies the "pseudo" nature of rain removal in this context and highlights the genuine need for snow removal. Then, aiming at the problems of snow removal and veil effect, it proposes a surveillance image restoration algorithm based on dual-tree complex wavelet transform, and develops two modules—the DTCWE module and the DCA module. Moreover, with a thorough understanding of the essence of the veil effect, it employs a RLR module to eliminate the veil effect in images. Finally, it demonstrates the positive impact of the modules on image restoration through visual examples and detailed explanations. It also presents the experimental design, loss function, platform setup, and evaluation metrics, with a thorough analysis of the results, confirming the algorithm's effective restoration of degraded surveillance images.

## REFERENCES

[1] S. K. Nayar and S. G. Narasimhan, "Vision in bad weather," presented at the *Seventh IEEE International Conference on Computer Vision*, Kerkyra, Greece, 1999, pp. 820-827 vol.2.

[2] R. Li, R. Tan, L. F. Cheong, A. Aviles-Rivero, Q. Fan and C. Schoenlieb, "RainFlow: Optical Flow Under Rain Streaks and Rain Veiling Effect," presented at the *2019 IEEE/CVF International Conference on Computer Vision (ICCV)*, Seoul, Korea (South), 2019, pp. 7303-7312.

[3] J. Redmon, S. Divvala, R. Girshick and A. Farhadi, "You Only Look Once: Unified, Real-Time Object Detection," presented at the *2016 IEEE Conference on Computer Vision and Pattern Recognition (CVPR)*, Las Vegas, NV, USA, 2016, pp. 779-788.

[4] W. Yang, R. T. Tan, J. Feng, J. Liu, Z. Guo and S. Yan, "Deep Joint Rain Detection and Removal from a Single Image," presented at the *2017 IEEE Conference on Computer Vision and Pattern Recognition (CVPR)*, Honolulu, HI, USA, 2017, pp. 1685-1694.

[5] A. W. M. Smeulders, D. M. Chu, R. Cucchiara, S. Calderara, A. Dehghan and M. Shah, "Visual Tracking: An Experimental Survey," *IEEE Transactions on Pattern Analysis and Machine Intelligence*, vol. 36, no. 7, pp. 1442-1468, July 2014, doi: 10.1109/TPAMI.2013.230.

[6] H. Mokayed, A. Nayebiastaneh, K. De, S. Sozos, O. Hagner and B. Backe, "Nordic Vehicle Dataset (NVD): Performance of vehicle detection using newly captured NVD from UAV in different snowy weather conditions," presented at the *2023 IEEE/CVF Conference on Computer Vision and Pattern Recognition Workshops (CVPRW)*, Vancouver, BC, Canada, 2023, pp. 5314-5322.

[7] B. Cai, X. Xu, K. Jia, C. Qing and D. Tao, "DehazeNet: An End-to-End System for Single Image Haze Removal," *IEEE Transactions on Image Processing*, vol. 25, no. 11, pp. 5187-5198, Nov. 2016, doi: 10.1109/TIP.2016.2598681.

[8] Y. -F. Liu, D. -W. Jaw, S. -C. Huang and J. -N. Hwang, "DesnowNet: Context-Aware Deep Network for Snow Removal," *IEEE*

*Transactions on Image Processing*, vol. 27, no. 6, pp. 3064-3073, June 2018, doi: 10.1109/TIP.2018.2806202.

[9]   W. T. Chen, H. Y. Fang, J. J. Ding, "JSTASR: Joint Size and Transparency-Aware Snow Removal Algorithm Based on Modified Partial Convolution and Veiling Effect Removal", *Cham*, 2020.DOI:10.1007/978-3-030-58589-1_45.

[10]  R. Li, R. T. Tan and L. -F. Cheong, "All in One Bad Weather Removal Using Architectural Search," presented at the *2020 IEEE/CVF Conference on Computer Vision and Pattern Recognition (CVPR)*, Seattle, WA, USA, 2020, pp. 3172-3182, doi: 10.1109/CVPR42600.2020.00324.

[11]  B. Li, X. Liu, P. Hu, Z. Wu, J. Lv and X. Peng, "All-In-One Image Restoration for Unknown Corruption," presented at the *2022 IEEE/CVF Conference on Computer Vision and Pattern Recognition (CVPR)*, New Orleans, LA, USA, 2022, pp. 17431-17441, doi: 10.1109/CVPR52688.2022.01693.

[12]  H. H. Wang, F. J. Tsai, Y. Y. Lin, "TANet: Triplet Attention Network forAll-In-One Adverse Weather Image Restoration," presented at the *2025 Asian Conference on Computer Vision*, Singapore, doi:10.1007/978-981-96-0911-61.

[13]  C. Wang, Z. Zheng, R. Quan, Y. Sun and Y. Yang, "Context-Aware Pretraining for Efficient Blind Image Decomposition," presented at the *2023 IEEE/CVF Conference on Computer Vision and Pattern Recognition (CVPR)*, Vancouver, BC, Canada, 2023, pp. 18186-18195, doi: 10.1109/CVPR52729.2023.01744.

[14]  V. Potlapalli, S. W. Zamir, S. H. Khan, "Promptir: Prompting for all-in-one image restoration," *Advances in Neural Information Processing Systems*, 2024, 36.

[15]  D. Fan, J. Zhang, L. Chang. "ConStyle v2: A Strong Prompter for All-in-One Image Restoration," 2024, *arXiv:2406.18242*.

[16]  X. Yu, S. Zhou, H. Li and L. Zhu, "Multi-Expert Adaptive Selection: Task-Balancing for All-in-One Image Restoration," *IEEE Transactions on Circuits and Systems for Video Technology*, doi: 10.1109/TCSVT.2024.3516074.

[17]  Q. Yan, A. Jiang, K. Chen, et al. "Textual prompt guided image restoration," 2023, *arXiv:2312.06162*.

[18]  Z, Li, Y, Lei, C, Ma, et al. "Prompt-in-prompt learning for universal image restoration," 2023, *arXiv:2312.05038*.

[19]  Y, Tian, J, Han, Chen H, et al. "Instruct-IPT: All-in-One Image Processing Transformer via Weight Modulation," 2024, *arXiv:2407.00676*.

[20]  W. T. Chen et al., "ALL Snow Removed: Single Image Desnowing Algorithm Using Hierarchical Dual-tree Complex Wavelet Representation and Contradict Channel Loss," presented at the *2021 IEEE/CVF International Conference on Computer Vision (ICCV)*, Montreal, QC, Canada, 2021, pp. 4176-4185, doi: 10.1109/ICCV48922.2021.00416.

[21]  J. M. Jose Valanarasu, R. Yasarla and V. M. Patel, "TransWeather: Transformer-based Restoration of Images Degraded by Adverse Weather Conditions," presented at the *2023 IEEE/CVF Conference on Computer Vision and Pattern Recognition (CVPR)*, New Orleans, LA, USA, 2022, pp. 2343-2353, doi: 10.1109/CVPR52688.2022.00239.

[22]  Y. Quan, X. Tan, Y. Huang, Y. Xu and H. Ji, "Image Desnowing via Deep Invertible Separation," *IEEE Transactions on Circuits and Systems for Video Technology*, vol. 33, no. 7, pp. 3133-3144, July 2023, doi: 10.1109/TCSVT.2022.3233655.

[23]  Y. Wang, C. Ma and J. Liu, "SmartAssign:Learning A Smart Knowledge Assignment Strategy for Deraining and Desnowing," presented at the *2023 IEEE/CVF Conference on Computer Vision and Pattern Recognition (CVPR)*, Vancouver, BC, Canada, 2023, pp. 3677-3686, doi: 10.1109/CVPR52729.2023.00358.

[24]  S. Chen et al., "MSP-Former: Multi-Scale Projection Transformer for Single Image Desnowing," presented at the *ICASSP 2023 - 2023 IEEE International Conference on Acoustics, Speech and Signal Processing (ICASSP)*, Rhodes Island, Greece, 2023, pp. 1-5, doi: 10.1109/ICASSP49357.2023.10095605.

[25]  Ting Zhang, Nanfeng Jiang, Hongxin Wu, Keke Zhang, Yuzhen Niu, and Tiesong Zhao. HCSD-Net: Single Image Desnowing with Color Space Transformation. presented at the *31st ACM International Conference on Multimedia (MM '23). Association for Computing Machinery*, New York, NY, USA, 2023, 8125–8133. doi: 10.1145/3581783.3613789.

[26]  D. Cho, T. D. Bui, "Multivariate statistical modeling for image denoising using wavelet transforms". *Signal Processing Image Communication*, 2005, 20(1): 77-89. doi: 10.1016/j.image.2004.10.003.

[27]  Y. Liu, "Image denoising method based on threshold, wavelet transform and genetic algorithm," *International Journal of Signal Processing, Image Processing and Pattern Recognition*, 2015, 8 (2), 29–40.

[28]  T. Guo, H. S. Mousavi, T. H. Vu and V. Monga, "Deep Wavelet Prediction for Image Super-Resolution," presented at the *2017 IEEE Conference on Computer Vision and Pattern Recognition Workshops (CVPRW)*, Honolulu, HI, USA, 2017, pp. 1100-1109, doi: 10.1109/CVPRW.2017.148.

[29]  H. Yang and Y. Wang, "An Effective and Comprehensive Image Super Resolution Algorithm Combined With a Novel Convolutional Neural Network and Wavelet Transform," *IEEE Access*, vol. 9, pp. 98790-98799, 2021, doi: 10.1109/ACCESS.2021.3083577.

[30]  A. Singh and N. Kingsbury, "Efficient convolutional network learning using parametric log based dual-tree wavelet scatternet," presented at the *2017 IEEE International Conference on Computer Vision Workshops*, 2017, pp. 1140–1147.

[31]  D. Li, L. Zhang, C. Sun, T. Yin, C. Liu, and J. Yang, "Robust retinal image enhancement using dual-tree complex wavelet transform and morphology-based method," *IEEE Access*, vol. 7, pp. 47 303–47 316, 2019. 3

[32]  H. Lu, H. Wang, Q. Zhang, D. Won, and S. W. Yoon, "A dualtree complex wavelet transform based convolutional neural network for human thyroid medical image segmentation," presented at the *2018 IEEE International Conference on Healthcare Informatics (ICHI)*, 2018, pp. 191–198. 3

[33]  H. Chen and N. Kingsbury, "Efficient registration of nonrigid 3-d bodies," *IEEE transactions on image processing*, vol. 21, no. 1, pp. 262–272, 2011. 3

[34]  T. Sun and C. Jung, "Readability enhancement of low light images based on dual-tree complex wavelet transform," presented at the *2016 IEEE International Conference on Acoustics, Speech and Signal Processing (ICASSP)*. IEEE, 2016, pp. 17411745. 3

[35]  C. Jung, Q. Yang, T. Sun, Q. Fu, and H. Song, "Low light image enhancement with dual-tree complex wavelet transform," *Journal of Visual Communication and Image Representation*, vol. 42, pp. 28–36, 2017.

[36]  Q. M. Tran, T. D. Ngo and T. -D. Mai, "Contrastive Learning with Weakly Pair Images for Traffic Image Deraining," presented at the *2024 International Conference on Multimedia Analysis and Pattern Recognition (MAPR)*, Da Nang, Vietnam, 2024, pp. 1-6, doi: 10.1109/MAPR63514.2024.10660776.

[37]  P. S. Baiju and S. N. George, "An Automated Unified Framework for Video Deraining and Simultaneous Moving Object Detection in Surveillance Environments," *IEEE Access*, vol. 8, pp. 128961-128972, 2020, doi: 10.1109/ACCESS.2020.3008903.

[38]  M. R. Islam and M. Paul, "Video Deraining Using the Visual Properties of Rain Streaks," *IEEE Access*, vol. 10, pp. 202-212, 2022, doi: 10.1109/ACCESS.2021.3136551.

[39]  V. M. Galshetwar, A. Kulkarni and S. Chaudhary, "Consolidated Adversarial Network for Video De-raining and De-hazing," presented at the *2022 18th IEEE International Conference on Advanced Video and Signal Based Surveillance (AVSS)*, Madrid, Spain, 2022, pp. 1-8, doi: 10.1109/AVSS56176.2022.9959454.

[40]  H. Gupta, O. Kotlyar, H. Andreasson and A. J. Lilienthal, "Robust Object Detection in Challenging Weather Conditions," presented at the *2024 IEEE/CVF Winter Conference on Applications of Computer Vision (WACV)*, Waikoloa, HI, USA, 2024, pp. 7508-7517, doi: 10.1109/WACV57701.2024.00735.

[41]  I. W. Selesnick, R. G. Baraniuk, and N. C. Kingsbury, "The dual-tree complex wavelet transform," *IEEE signal processing magazine*, vol. 22, no. 6, pp. 123–151, 2005. 2, 3.

[42]  P. Hill, M. E. Al-Mualla and D. Bull, "Perceptual Image Fusion Using Wavelets," *IEEE Transactions on Image Processing*, vol. 26, no. 3, pp. 1076-1088, March 2017, doi: 10.1109/TIP.2016.2633863.

[43]  P. Hill, A. Achim, M. E. Al-Mualla and D. Bull, "Contrast Sensitivity of the Wavelet, Dual Tree Complex Wavelet, Curvelet, and Steerable Pyramid Transforms," *IEEE Transactions on Image Processing*, vol.

25, no. 6, pp. 2739-2751, June 2016, doi: 10.1109/TIP.2016.2552725.
k.

[44] D. Li, L. Zhang, C. Sun, T. Yin, C. Liu and J. Yang, "Robust Retinal Image Enhancement via Dual-Tree Complex Wavelet Transform and Morphology-Based Method," *IEEE Access*, vol. 7, pp. 47303-47316, 2019, doi: 10.1109/ACCESS.2019.2909788.

[45] M. Fierro, H. -G. Ha and Y. -H. Ha, "Noise Reduction Based on Partial-Reference, Dual-Tree Complex Wavelet Transform Shrinkage," *IEEE Transactions on Image Processing*, vol. 22, no. 5, pp. 1859-1872, May 2013, doi: 10.1109/TIP.2013.2237918.

[46] Y. Chen, X. Dai, M. Liu, D. Chen, L. Yuan and Z. Liu, "Dynamic Convolution: Attention Over Convolution Kernels," presented at the *2020 IEEE/CVF Conference on Computer Vision and Pattern Recognition (CVPR)*, Seattle, WA, USA, 2020, pp. 11027-11036, doi: 10.1109/CVPR42600.2020.01104.

[47] W. Bae, J. Yoo and J. C. Ye, "Beyond Deep Residual Learning for Image Restoration: Persistent Homology-Guided Manifold Simplification," presented at the *2017 IEEE Conference on Computer Vision and Pattern Recognition Workshops (CVPRW)*, Honolulu, HI, USA, 2017, pp. 1141-1149, doi: 10.1109/CVPRW.2017.152.

[48] Y. Zhang, Y. Tian, Y. Kong, B. Zhong and Y. Fu, "Residual Dense Network for Image Restoration," *IEEE Transactions on Pattern Analysis and Machine Intelligence*, vol. 43, no. 7, pp. 2480-2495, 1 July 2021, doi: 10.1109/TPAMI.2020.2968521.

[49] K. He, J. Sun and X. Tang, "Single Image Haze Removal Using Dark Channel Prior," *IEEE Transactions on Pattern Analysis and Machine Intelligence*, vol. 33, no. 12, pp. 2341-2353, Dec. 2011, doi: 10.1109/TPAMI.2010.168.

[50] Z. Zhang, "Improved Adam Optimizer for Deep Neural Networks," presented at the *2018 IEEE/ACM 26th International Symposium on Quality of Service (IWQoS)*, Banff, AB, Canada, 2018, pp. 1-2, doi: 10.1109/IWQoS.2018.8624183.

[51] Junjun Jiang and Zengyuan Zuo and Gang Wu and Kui Jiang and Xianming Liu. "A Survey on All-in-One Image Restoration: Taxonomy, Evaluation and Future Trends," 2024, arXiv: 2410.15067.

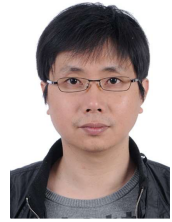

**QINGYU YANG** received Master's degree in at the School of Electrical Engineering, Chongqing University. He is currently pursuing the Ph.D in Electronic and Information Engineering at Tianjin University, and he is also a professorate senior engineer at the Automatic Control Research Institute of China Nuclear Power Engineering Co., Ltd. His research focuses on the design of intelligent main control rooms for nuclear power plants, advanced control systems and artificial intelligence.

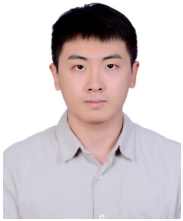

**ZIHAN SHEN** received Master's degree in Electronic and Information Engineering from the University of Chinese Academy of Sciences. He is currently an engineer at the Automatic Control Research Institute of China Nuclear Power Engineering Co., Ltd. His research interests include exploring emerging technologies such as machine learning, computer vision, and modern control systems. Additionally, he is passionate about applying artificial intelligence to solve engineering challenges.

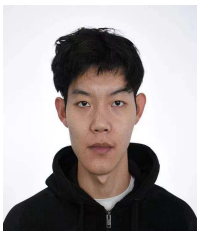

**YU XUAN** received Master's degree in Digital Image and Signal Processing from the University of California San Diego. He is currently an engineer at Chinese aeronautical establishment of Aviation Industry Corporation of China, Ltd. His research interests include exploring emerging technologies such as machine learning, image processing.